\documentclass[letterpaper, 10 pt, conference]{ieeeconf}  

\IEEEoverridecommandlockouts    

\usepackage{graphics} 
\usepackage{cite} 
\usepackage{mathrsfs}
\usepackage{times}
\usepackage{epsfig}
\usepackage{amsmath}
\usepackage{amssymb}
\usepackage{threeparttable}
\usepackage{booktabs}
\usepackage{amsfonts, amssymb}
\usepackage{multirow,bigdelim}
\usepackage{subfigure}
\usepackage{indentfirst}
\usepackage{caption}
\usepackage{xcolor}
\usepackage{xcolor,colortbl}

\usepackage[pagebackref=true,breaklinks=true,letterpaper=true, colorlinks,bookmarks=false]{hyperref}

\definecolor{car}{rgb}{0.39215686, 0.58823529, 0.96078431}
\definecolor{bicycle}{rgb}{0.39215686, 0.90196078, 0.96078431}
\definecolor{motorcycle}{rgb}{0.11764706, 0.23529412, 0.58823529}
\definecolor{truck}{rgb}{0.31372549, 0.11764706, 0.70588235}
\definecolor{other-vehicle}{rgb}{0.39215686, 0.31372549, 0.98039216}
\definecolor{person}{rgb}{1.        , 0.11764706, 0.11764706}
\definecolor{bicyclist}{rgb}{1.        , 0.15686275, 0.78431373}
\definecolor{motorcyclist}{rgb}{0.58823529, 0.11764706, 0.35294118}
\definecolor{road}{rgb}{1.        , 0.        , 1.        }
\definecolor{parking}{rgb}{1.        , 0.58823529, 1.        }
\definecolor{sidewalk}{rgb}{0.29411765, 0.        , 0.29411765}
\definecolor{other-ground}{rgb}{0.68627451, 0.        , 0.29411765}
\definecolor{building}{rgb}{1.        , 0.78431373, 0.        }
\definecolor{fence}{rgb}{1.        , 0.47058824, 0.19607843}
\definecolor{vegetation}{rgb}{0.        , 0.68627451, 0.        }
\definecolor{trunk}{rgb}{0.52941176, 0.23529412, 0.        }
\definecolor{terrain}{rgb}{0.58823529, 0.94117647, 0.31372549}
\definecolor{pole}{rgb}{1.        , 0.94117647, 0.58823529}
\definecolor{traffic-sign}{rgb}{1.        , 0.        , 0.    }

\title{\LARGE \bf
Semantic Segmentation-assisted Scene Completion \\
for LiDAR Point Clouds
}

\author{Xuemeng Yang$^{1}$, Hao Zou$^{1}$, Xin Kong$^{1}$, Tianxin Huang$^{1}$, Yong Liu$^{1,*}$, \\
Wanlong Li$^{2}$, Feng Wen$^{2}$, and Hongbo Zhang$^{2}$
\thanks{$^{1}$The authors are with the Institute of Cyber-Systems and Control, Zhejiang University, Hangzhou, 310027, China. (Yong Liu* is the corresponding author, email: yongliu@iipc.zju.edu.cn)}
\thanks{$^{2}$The authors are with Huawei Noah’s Ark Lab, Beijing, China.}
}
\begin{document}

\maketitle
\thispagestyle{empty}
\pagestyle{empty}

\begin{abstract}

	Outdoor scene completion is a challenging issue in 3D scene understanding, which plays an important role in intelligent robotics and autonomous driving. Due to the sparsity of LiDAR acquisition, it is far more complex for 3D scene completion and semantic segmentation. Since semantic features can provide constraints and semantic priors for completion tasks, the relationship between them is worth exploring. Therefore, we propose an end-to-end semantic segmentation-assisted scene completion network, including a 2D completion branch and a 3D semantic segmentation branch. Specifically, the network takes a raw point cloud as input, and merges the features from the segmentation branch into the completion branch hierarchically to provide semantic information. By adopting BEV representation and 3D sparse convolution, we can benefit from the lower operand while maintaining effective expression. Besides, the decoder of the segmentation branch is used as an auxiliary, which can be discarded in the inference stage to save computational consumption. Extensive experiments demonstrate that our method achieves competitive performance on SemanticKITTI dataset with low latency. Code and models will be released at \href{https://github.com/jokester-zzz/SSA-SC}{https://github.com/jokester-zzz/SSA-SC}.

\end{abstract}

\section{Introduction}

\begin{figure}[t]
	\centering
	\includegraphics[width=8.5cm]{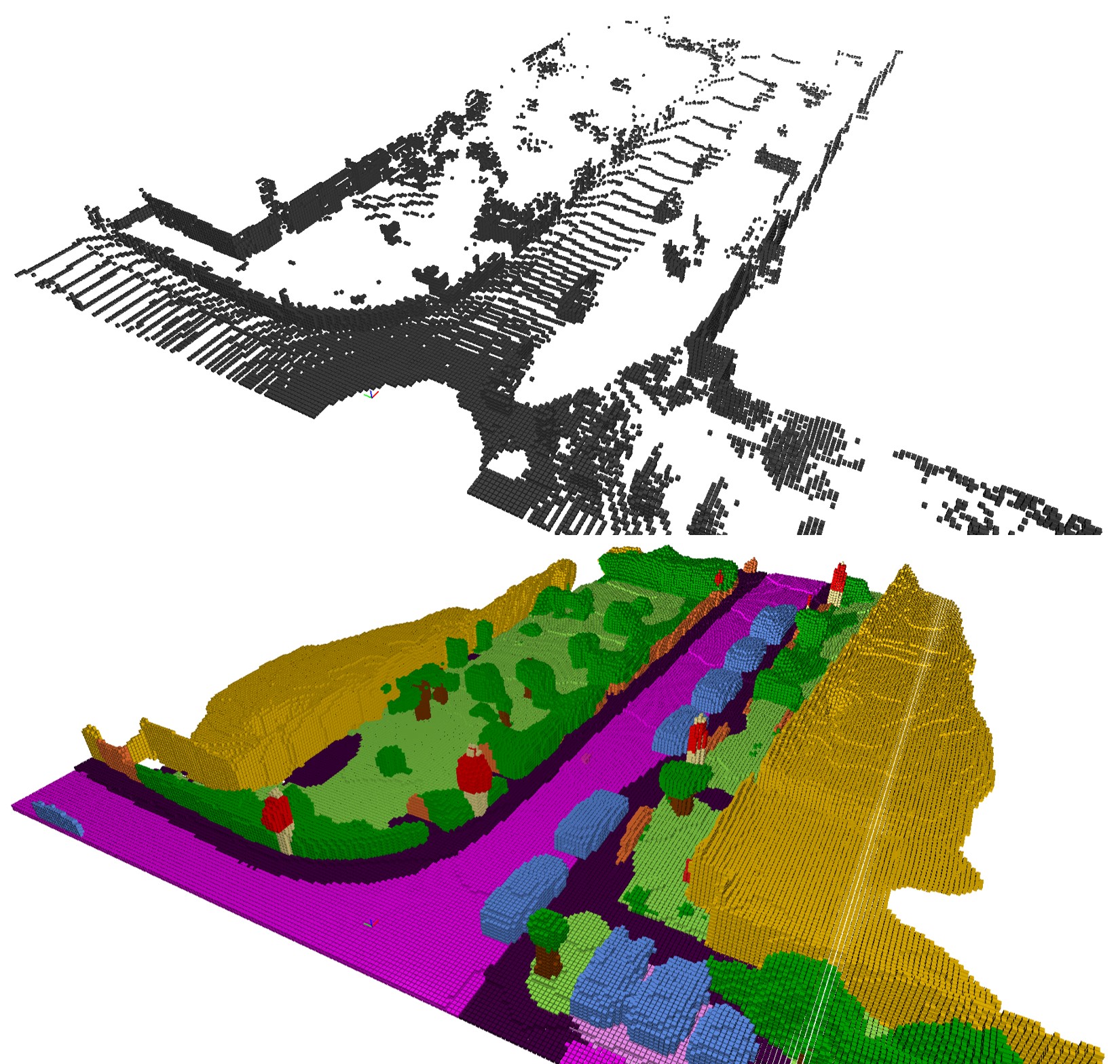}
	\caption{Semantic scene completion on SemanticKITTI dataset. The input is the sparse raw point cloud and the output is the complete semantic scene predicted by our approach.}
	\label{fig:introduction}
 \end{figure}

Scene completion plays an important role in autonomous driving, which is a fundamental block of 3D scene understanding. In autonomous driving scenarios, LiDAR is the most commonly used 3D sensor. However, the point cloud collected by a LiDAR is inherently sparse due to its acquisition method, and can only collect data on object surface, which makes it more difficult for machines to infer and understand the scene than humans. Therefore, semantic scene completion task has been proposed to complete the entire scene from limited information, and segment the semantics. Fig.~\ref{fig:introduction} is an illustration of outdoor point cloud scene completion. 

Compared with scene completion of depth image, the point cloud scene completion contains much more input data, which will bring greater challenges. Furthermore, since the memory will grows cubically with the increasing with the input voxel resolution, the 3D convolutional neural network also requires noticeable overhead costs~\cite{song2017semantic}. 
It was not until the emergence of sparse convolution~\cite{graham20183d} that this situation was alleviated. However, submanifold sparse convolution always maintains the sparsity of voxels, it is hard to be directly applied to the task of scene completion. So Zhang et al.~\cite{zhang2018efficient} insert a ``dense'' deconvolution layer in the Sparse Convolutional Network (SCN) to generate new voxels. LMSCNet~\cite{roldao2020lmscnet} tackles this problem by using a lightweight 2D convolutions followed by a 3D segmentation head block. 

Inspired by~\cite{yan2020sparse} which adopts a completion network after semantic segmentation, we explore the relationship between semantic segmentation and scene completion tasks. For scene completion, semantic features are critical in recovering the complete shape of the specific objects and the entire scene. It's easy to infer the whole shape from an incomplete object by identifying it's category, which means the semantics essentially provide the constraints and priori for completion. Based on this insight, we believe that semantic segmentation tasks and scene completion tasks are inseparable, and semantic segmentation can provide potential semantic information for the scene completion task.

In this paper, we try to explore the combination of semantic segmentation and scene completion, and benefit from both the 2D and 3D convolutional neural network. The whole network consists of two parts including a 2D completion branch and an assistant 3D segmentation branch. Since 3D dense convolution consumes too much resources, and 3D sparse convolution is difficult to generate new voxels, we use a 2D network on Bird's Eye View (BEV) to complete the scene. It is efficient and easy to diffuse features with 2D convolution. In addition, we introduce the features of the semantic segmentation branch as an auxiliary. We assume features from the semantic segmentation branch as source that can continuously deliver semantic features to the completion branch with combining the advantages of 2D and 3D networks from multi-view fusion. The main contributions of this paper are three-fold:
\begin{itemize}
   \item We propose a novel semantic segmentation-assisted scene completion network that leverages the complementary information between BEV map and 3D voxels from multi-views.
   \item The proposed network can produce reasonable completion results with low latency and memory cost for outdoor 3D scenes by the aid of auxiliary semantic segmentation branch.
   \item Experiments on SemanticKITTI dataset show that our approach achieves the state-of-the-art performance. We rank the $3^{rd}$ in public semantic scene completion benchmark\footnote{\url{https://competitions.codalab.org/competitions/22037\#results}} and outperform all the published work in terms of completion metric (IoU).
   
\end{itemize}

\section{Related Work}

\begin{figure*}[t]
   \centering
   \includegraphics[width=18cm]{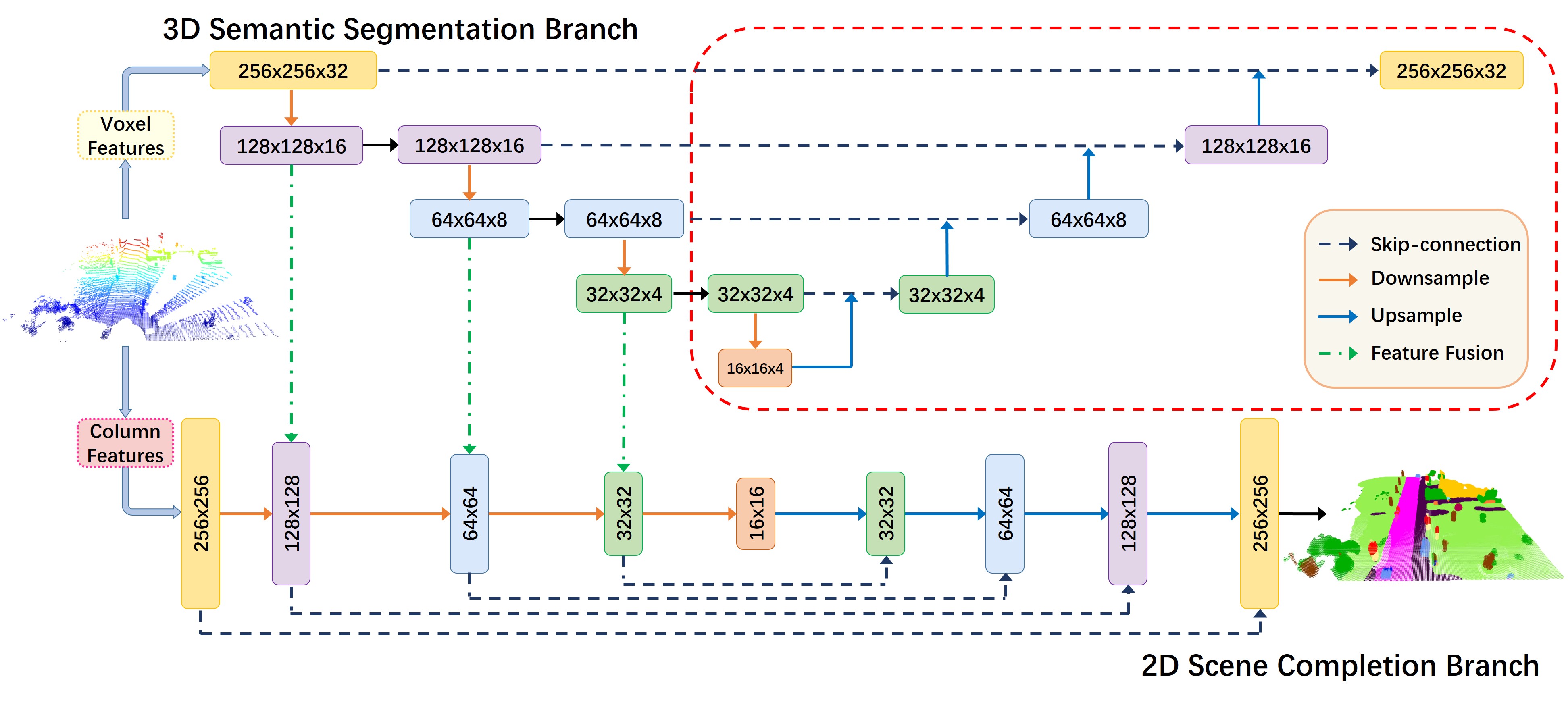}
   \caption{The network structure of the proposed method. The upper part of the figure is an auxiliary 3D semantic segmentation branch, and the lower part is a 2D completion branch. Both branches follow the UNet structure and carry out four downsamplings. We merge the first three downsampling results from the semantic segmentation branch into the completion branch at the same level to supplement semantic information. The same color rectangles represent the same downsampling stage, and the numbers in the rectangles represent the feature resolution. In the inference stage, the part in the red dashed line can be discarded to further save memory. (Best viewed in color.)}
   \label{fig:backbone}
   \vspace{-0.3cm}
\end{figure*}

\begin{figure}[h]
	\centering
	\includegraphics[width=9cm]{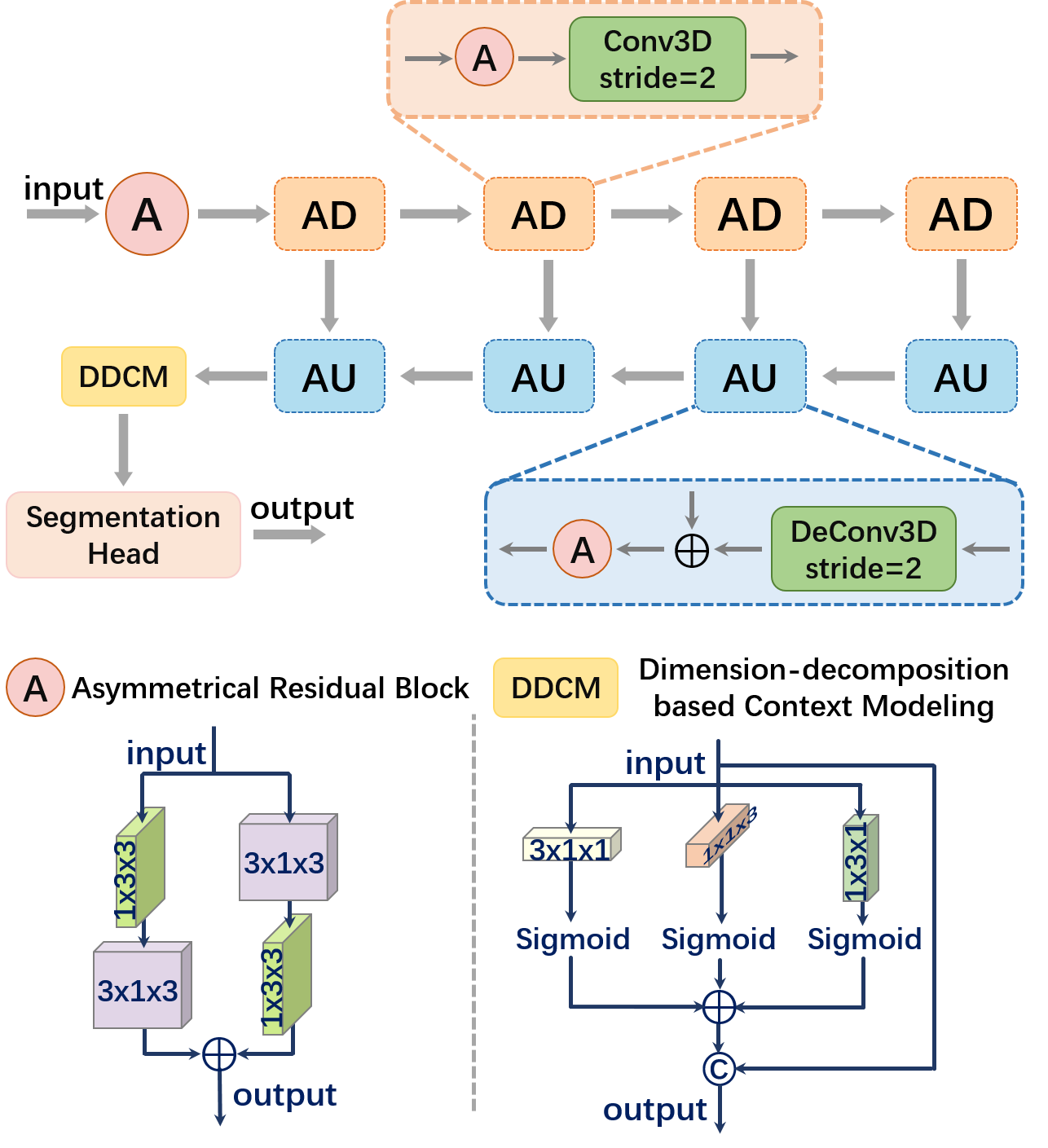}
	\caption{The structure of the assistant semantic segmentation network, where AD stands for asymmetrical downsample block and AU stands for asymmetrical upsample block.}
	\label{fig:3DUnet}
 \end{figure}

In this section, we introduce the development of point cloud segmentation and completion.

\subsection{3D semantic segmentation}

Semantic segmentation on point clouds can be classified into three types: projection-based, voxel-based and point-based. 

Projection-based works usually project raw point clouds onto various 2D planes. SqueezeSegs~\cite{wu2017squeezeseg, wu2018squeezesegv2, xu2020squeezesegv3} and RangeNet++~\cite{milioto2019iros} project the point cloud onto the spherical image. SalsaNet~\cite{aksoy2019salsanet} organizes point clouds into BEV feature maps. PolarNet~\cite{Zhang_2020_CVPR} proposes a polar BEV representation, achieving a balanced grid distribution. They all use 2D semantic segmentation networks as the backbone. 
Voxel-based methods split the raw point clouds into tightly arranged voxels and usually apply 3D convolutions to abstract features. Due to the sparsity, 3D sparse convolution~\cite{graham20183d} is widely used and 3D UNet-like~\cite{cciccek20163d} is often adopted as the backbone network for semantic segmentation. Cylinder3D~\cite{zhou2020cylinder3d} and Cylinder3D++~\cite{zhu2020cylindrical} use a cylindrical partition on the point cloud and design an Asymmetric Residual Block and a Dimension-decomposition based Context Modeling based on 3D sparse convolution, which can significantly reduce the computational cost. Zhang et al.~\cite{zhang2020deep} propose a deep fusion network architecture with a unique voxel-based “mini-PointNet” point cloud representation, which fully integrates the features of points and voxels.
Some researchers continue the PointNets~\cite{qi2017pointnet,qi2017pointnet++} approach, processing on the original point cloud without intermediate representation. They aim to segment the point cloud by directly extracting features from 3D points hierarchically to capture the local or global context~\cite{hu2020randla}, or design a new convolution method~\cite{xu2018spidercnn, thomas2019kpconv}.

Among all these methods, the voxel-based method can be best combined with the scene completion task, as it is convenient to indicate whether the space is occupied or not.

\subsection{3D scene completion}

Given a single frame point cloud or depth map as input, the scene completion work should predict $C+1$ labels for each voxel in the 3D space, indicating whether it is occupied and what the semantic category is. The earliest scene completion work is based on depth image proposed by Song et al.~\cite{song2017semantic}. They formulate an end-to-end 3D ConvNet model (SSCNet) for the volumetric scene completion and semantic labeling with a dilation-based 3D context module. In order to complete the scene, the entire network is composed of dense 3D convolution, which consumes a lot of resources. 
Most of the subsequent works are based on 3D convolution networks. 
ScanComplete~\cite{dai2018scancomplete} introduces a coarse-to-fine inference strategy to produce high-resolution output. SGC~\cite{zhang2018efficient} benefits from 3D sparse convolution and adopts Spatial Group Convolutions for efficient processing at the cost of small performance degradation. SATNet~\cite{liu2018see} decomposes semantic scene completion tasks into 2D semantic segmentation and 3D scene completion, which are connected by a 2D-3D reprojection layer. 

Until the emergence of the SemanticKITTI dataset~\cite{behley2019semantickitti}, researchers turn their attention to the semantic scene completion of outdoor point clouds. LMSCNet~\cite{roldao2020lmscnet} designs a lightweight network with a mix of 2D and 3D convolutions, using a 2D UNet backbone followed by a 3D segmentation head. In the case of using only occupied voxels as input, they achieve acceptable results. Rist et al.~\cite{rist2020semantic} produce a continuous scene representation instead of voxelization. This method can predict any position in the scene without the need for spatial discretization. S3CNet~\cite{cheng2020s3cnet} proposes a multi-view fusion method that performs semantic completion of the scene in 2D and 3D respectively, and finally uses a dynamic voxel fusion to merge the results. Besides, they leverage LiDAR-based flipped truncated symbol distance function (fTSDF~\cite{song2017semantic}) calculated from the spherical range image and point-wise normal vectors as spatial encoding. JS3C-Net~\cite{yan2020sparse} adds a SSCNet after semantic segmentation network for completion, and proposes a Point-Voxel Interaction (PVI) module for refinement. However, it cannot perform in real-time due to the cascade architecture.

In this paper, we propose an end-to-end network that completely uses the network to extract features without manually designed features. We design a 2D scene completion network assisted by a semantic segmentation branch, whose decoder part is discarded in the inference stage to achieve a fast speed while reserving a helpful encoder with 3D sparse convolution layers.

\section{Methodology}

The problem we solve is to infer the entire scene from a single frame point cloud. In this work, we combine 2D and 3D convolutional networks, as illustrated in Fig.~\ref{fig:backbone}. We use a 2D network for scene completion since it is far less complex and a 3D segmentation network as an auxiliary branch for the completion branch. The input of the network is a point cloud $P$ with points features $f_p$, which represents the coordinates of points and their corresponding point-wise information. The output is a label of $C+1$ categories for each voxel, where $C$ is the number of semantic categories, indicating whether a voxel is free or occupied by a semantic category.

\subsection{Scene Completion Branch} 
\label{sec:3.1}
Since the 3D completion network often needs ``dense'' convolution for dilation, it consumes more resources. In contrast, 2D networks are more lightweight and convenient in diffusing features. Therefore, we adopt a 2D encoder-decoder architecture as the backbone. In order to adapt to the scene completion task, we carry out the Cartesian voxelization instead of the commonly used spherical projection in segmentation. Given the point cloud $P \in \mathbb{R}^{N\times3}$ in the range of $[R_x, R_y, R_z]$, we voxelize the irregular points into voxels with a resolution of $L \times W \times H$. 

Here, we adopt a top-down view to generate a BEV feature map with the size of $L \times W$. Since the BEV map is very similar to image, 2D CNN can be directly applied on it. 
As with some detection tasks~\cite{lang2019pointpillars}, we fuse the features of each column along the z-axis. Specifically, we use a simple MLP to learn point features, followed by a max-pooling layer in each column to get the preliminary column feature. After a feature dimension reduction layer, we obtain the column feature.
\begin{equation}\label{equ:mlp}
   f_{x, y}=\mathcal{A}_1\left\{\underset{p \in V_{x,y}}{\operatorname{MAX}}\left(\mathcal{M}(f_p)\right)\right\}.
\end{equation}
$\mathcal{A}_1$ is a dimension reduction layer including a simple Linear layer followed by a ReLU function. $\mathcal{M}$ represents an MLP that only contains fully connected layers, batch normalization and ReLU. $V_{x,y}$ denotes the $(x, y)^{th}$ column. $f_{p}$ represents the feature of point $p$ and can be defined flexibly. In our implementation, the point feature $f_p$ is defined as:
\begin{equation}
    f_{p}=(\Delta x, \Delta y, \Delta z, x, y, z, r),
\end{equation}
$(\Delta x, \Delta y, \Delta z)$ is the coordinate difference between the point $p$ and the center of the voxel it locates while $r$ denotes the reflection intensity.
So far, we have obtained a BEV feature of size $C_f \times L \times W$ that can represent the entire scene, where $C_f$ is the feature dimension. Then, we input the obtained BEV features into the 2D completion branch.

Unlike LMSCNet~\cite{roldao2020lmscnet} using a 3D segmentation head, we directly output a tensor of size $L \times W$, which encodes the prediction of each voxel along the height of that location. By reshaping the output tensor, the prediction of each voxel in the scene can be acquired. In order to fully diffuse the features, we use a series of 2D convolution layers to expand the receptive field. The entire network is downsampled four times, and each time the resolution size is reduced by 2 shown in Fig.\ref{fig:backbone}. Skip connections with concatenation are used at the same time. However, aggregating features with a 2D network inevitably lacks some information along the z-axis, so we supplement additional features with the aid of the semantic segmentation branch.

\subsection{Semantic Segmentation Branch}

To supplement the height information lost in the 2D convolution, we introduce a 3D sparse convolutional branch. The sparse convolution can aggregate the voxel features effectively and continuously provide semantic features for the completion branch hierarchically. These voxel features from the segmentation branch help the completion task in extending voxels and predicting their semantic categories. 

Similar to the operations in \textit{Section~\ref{sec:3.1}}, the input of the branch is aggregated voxel features.  Based on the Cartesian voxelization, the point-wise features obtained from the same MLP in Eq.\ref{equ:mlp} are reassigned to obtain voxel features. With a  max pooling performed in each voxel and another dimension reduction layer $\mathcal{A}_2$ (a Linear layer followed by a ReLU), we obtain the input of the 3D branch.
\begin{equation}
   f_{x,y,z}=\mathcal{A}_2\left\{\underset{p \in V_{x,y,z}}{\operatorname{MAX}}\left(\mathcal{M}(f_p)\right)\right\}, 
\end{equation}
where $V_{x,y,z}$ denotes the $(x,y,z)^{th}$ voxel. Reusing the same MLP but only adopting different dimension reduction layers can save some resource consumption. Now, we obtain a feature of size $C_f \times L \times W \times H$ that represents the entire scene, where $C_f$ is the feature dimension and is consistent with Section~\ref{sec:3.1}. Then, we input the obtained features into the 3D segmentation branch.

To better supplement features for the completion branch with each corresponding layer, we choose the same encoder-decoder structure in this semantic segmentation branch modified from the 3D UNet in Cylinder3D~\cite{zhou2020cylinder3d}, which is implemented with 3D sparse convolution.
As shown in Fig.~\ref{fig:3DUnet}, both the asymmetrical downsample and upsample blocks in the network contain Asymmetrical Residual Block. The usage of the convolution with a $3\times 1\times 3$ kernel followed by a $1\times 3\times 3$ kernel is equivalent to sliding a two-layer network with a conventional $3\times 3\times 3$ kernel, but can save 33\% of memory consumption. The asymmetrical downsample block also halves the resolution size each time, so that the tensors output from the same level of the 3D and 2D network have the same size in terms of length and width. 
Dimension-decomposition based Context Modeling Block divides high-level context information into low-level features in three dimensions (length, width, height), and aggregates all three low-level activations to obtain features representing a complete context. We use the most frequently appeared semantic label in the raw occupied voxel as supervision.

We draw semantic features from the encoder part and input them into the completion branch. Specifically, we stack the 3D features along the z-axis and reduce the stacked feature dimension so that it can have the same dimension $(C_s \times L \times W)$ as the 2D feature of the same resolution. Then, we concatenate it with the completion feature in the $C_s$ dimension, which can provide features with information along the z-axis for the completion task. It is worth noting that we only use features from the encoder part and the decoder part can be discarded to further reduce the burden of GPU memory and calculation in the inference stage.

\subsection{Loss}
In the experiment, both lovasz loss~\cite{berman2018lovasz} and cross-entropy loss are used for the two branches. Lovasz loss is a method for directly optimizing the mean intersection-over-union (mIoU) metric, which is defined as: 
\begin{equation}\label{equ:lovasz}
	Loss_{lovasz}=\frac{1}{|C|} \sum_{c \in C} J(e(c)),
\end{equation}
where $J$ is the lovasz extension of IoU and denotes a piece-wise linear function with a global minimum, and $e(c)$ is the vector of errors for class $c$. Cross-entropy loss is widely used and optimizes the accuracy:
\begin{equation}
    Loss_{CE}=-\sum_{i} y_i \log \hat{y}_{i}.
\end{equation}
$\hat{y}_{i}$ and $y_i$ are the corresponding predicted and ground truth probability. The total loss is:
\begin{equation}
   Loss_{all} = \sigma_1 Loss_{seg} + \sigma_2 Loss_{com}.
\end{equation}
In our work, we set $\sigma_1=\sigma_2=0.5$, and the loss for two branches are:
\begin{equation}
   \begin{aligned}
      &Loss_{seg} = Loss_{lovasz} + Loss_{CE},\\
      &Loss_{com} = Loss_{lovasz} + Loss_{CE}.
   \end{aligned}
\end{equation}
In the ablation study, we verify the effectiveness of lovasz loss in the completion branch.

\begin{table*}
	\scriptsize
	\setlength{\tabcolsep}{0.005\linewidth}
	\centering
	\caption{Comparison of published methods on the official SemanticKITTI~\cite{behley2019semantickitti} benchmark. Our network surpasses all the published methods in terms of completion metrics (IoU), and ranks $3^{rd}$ on the semantic segmentation metrics (mIoU). (* originate from~\cite{roldao2020lmscnet}. The last column of data comes from their paper, the data in brackets are reproduced on our device.)} 
	\begin{tabular}{l|c|c c c c c c c c c c c c c c c c c c c|c c}
		\toprule
		Approach 
		& \rotatebox{0}{IoU}
		& \rotatebox{90}{\textcolor{road}{$\blacksquare$} road} 
		& \rotatebox{90}{\textcolor{sidewalk}{$\blacksquare$} sidewalk}
		& \rotatebox{90}{\textcolor{parking}{$\blacksquare$} parking} 
		& \rotatebox{90}{\textcolor{other-ground}{$\blacksquare$} other-ground} 
		& \rotatebox{90}{\textcolor{building}{$\blacksquare$} building} 
		& \rotatebox{90}{\textcolor{car}{$\blacksquare$} car} 
		& \rotatebox{90}{\textcolor{truck}{$\blacksquare$} truck} 
		& \rotatebox{90}{\textcolor{bicycle}{$\blacksquare$} bicycle} 
		& \rotatebox{90}{\textcolor{motorcycle}{$\blacksquare$} motorcycle} 
		& \rotatebox{90}{\textcolor{other-vehicle}{$\blacksquare$} other-vehicles} 
		& \rotatebox{90}{\textcolor{vegetation}{$\blacksquare$} vegetation} 
		& \rotatebox{90}{\textcolor{trunk}{$\blacksquare$} trunk} 
		& \rotatebox{90}{\textcolor{terrain}{$\blacksquare$} terrain} 
		& \rotatebox{90}{\textcolor{person}{$\blacksquare$} person} 
		& \rotatebox{90}{\textcolor{bicyclist}{$\blacksquare$} bicyclist}
		& \rotatebox{90}{\textcolor{motorcyclist}{$\blacksquare$} motorcyclist} 
		& \rotatebox{90}{\textcolor{fence}{$\blacksquare$} fence} 
		& \rotatebox{90}{\textcolor{pole}{$\blacksquare$} pole} 
		& \rotatebox{90}{\textcolor{traffic-sign}{$\blacksquare$} traffic-sign} 
		& \rotatebox{0}{mIoU} 
		& \rotatebox{0}{FPS} \\
		\midrule
		*SSCNet~\cite{song2017semantic} & 29.8 & 27.6 & 17.0 & 15.6 & 6.0 & 20.9 & 10.4 & 1.8 & 0.0 & 0.0 & 0.1 & 25.8 & 11.9 & 18.2 & 0.0 & 0.0 & 0.0 & 14.4 & 7.9 & 3.7 & 9.5 & \textbf{56.90} \\ %
		*SSCNet-full~\cite{song2017semantic} & 50.0 & 51.2 & 30.8 & 27.1 & 6.4 & 34.5 & 24.3 & 1.2 & 0.5 & 0.8 & 4.3 & 35.3 & 18.2 & 29.0 & 0.3 & 0.3 & 0.0 & 19.9 & 13.1 & 6.7 & 16.1 & 45.94 \\ %
		*TS3D~\cite{garbade2019two} & 29.8 & 28.0 & 17.0 & 15.7 & 4.9 & 23.2 & 10.7 & 2.4 & 0.0 & 0.0 & 0.2 & 24.7 & 12.5 & 18.3 & 0.0 & 0.1 & 0.0 & 13.2 & 7.0 & 3.5 & 9.5 & 9.79 \\ %
		*TS3D+DNet~\cite{behley2019semantickitti} & 25.0 & 27.5 & 18.5 & 18.9 & 6.6 & 22.1 & 8.0 & 2.2 & 0.1 & 0.0 & 4.0 & 19.5 & 12.9 & 20.2 & 2.3 & 0.6 & 0.0 & 15.8 & 7.6 & 7.0 & 10.2 & 8.72 \\ %
		*TS3D+DNet+SATNet~\cite{behley2019semantickitti} & 50.6 & 62.2 & 31.6 & 23.3 & 6.5 & 34.1 & 30.7 & 4.9 & 0.0 & 0.0 & 0.1 & 40.1 & 21.9 & 33.1 & 0.0 & 0.0 & 0.0 & 24.1 & 16.9 & 6.9 & 17.7 & 1.27\\ %
		
		LMSCNet~\cite{roldao2020lmscnet} & 55.3 & 64.0 & 33.1 & 24.9 & 3.2 & 38.7 & 29.5 & 2.5 & 0.0 & 0.0 & 0.1 & 40.5 & 19.0 & 30.8 & 0.0 & 0.0 & 0.0 & 20.5 & 15.7 & 0.5 & 17.0 & 21.28 (8.51) \\
		LMSCNet-singlescale~\cite{roldao2020lmscnet} & 56.7 & 64.8 & 34.7 & 29.0 & 4.6 & 38.1 & 30.9 & 1.5 & 0.0 & 0.0 & 0.8 & 41.3 & 19.9 & 32.1 & 0.0 & 0.0 & 0.0 & 21.3 & 15.0 & 0.8 & 17.6 & - \\
		Local-DIFs~\cite{rist2020semantic} & 57.7 & 67.9 & 42.9 & \textbf{40.1} & 11.4 & 40.4 & 34.8 & 4.4 & 3.6 & 2.4 & 4.8 & 42.2 & 26.5 & 39.1 & 2.5 & 1.1 & 0.0 & 29.0 & 21.3 & 17.5 & 22.7 & - \\
		JS3C-Net~\cite{yan2020sparse} & 56.6 & 64.7 & 39.9 & 34.9 & \textbf{14.1} & 39.4 & 33.3 & \textbf{7.2} & 14.4 & 8.8 & 12.7 & 43.1 & 19.6 & 40.5 & 8.0 & 5.1 & 0.4 & 30.4 & 18.9 & 15.9 & 23.8 & 1.73 (1.20) \\
		S3CNet~\cite{cheng2020s3cnet} & 45.6 & 42.0 & 22.5 & 17.0 & 7.9 & \textbf{52.2} & 31.2 & 6.7 & \textbf{41.5} & \textbf{45.0} & \textbf{16.1} & 39.5 & \textbf{34.0} & 21.2 & \textbf{45.9} & \textbf{35.8} & \textbf{16.0} & \textbf{31.3} & \textbf{31.0} & \textbf{24.3} & \textbf{29.5} & 1.82\\
		\midrule
		Ours & \textbf{58.8} & \textbf{72.2} & \textbf{43.7} & 37.4 & 10.9 & 43.6 & \textbf{36.5} & 5.7 & 13.9 & 4.6 & 7.4 & \textbf{43.5} & 25.6 & \textbf{41.8} & 4.4 & 2.6 & 0.7 & 30.7 & 14.5 & 6.9 & 23.5 & 20.04 \\
	  
		\bottomrule
	\end{tabular}\\
   
	\label{table:net_reults}
\end{table*}

\newcommand{\tabincell}[2]{\begin{tabular}{@{}#1@{}}#2\end{tabular}} 
\begin{table*}
	\scriptsize
	\setlength{\tabcolsep}{0.0035\linewidth}
	\centering
	\caption{Ablation experiment on SemanticKITTI dataset. The results are conducted on the validation set.} 
	\begin{tabular}{l c | c c|c c c|c c c c c c c c c c c c c c c c c c c c}
		\toprule
        2D & 3D & \tabincell{c}{Com \\ CE / lvz} & \tabincell{c}{Seg \\ CE / lvz}
		& \rotatebox{90}{IoU}
		& \rotatebox{90}{Precision}
		& \rotatebox{90}{Recall}
		& \rotatebox{90}{mIoU}
		& \rotatebox{90}{\textcolor{road}{$\blacksquare$} road} 
		& \rotatebox{90}{\textcolor{sidewalk}{$\blacksquare$} sidewalk}
		& \rotatebox{90}{\textcolor{parking}{$\blacksquare$} parking} 
		& \rotatebox{90}{\textcolor{other-ground}{$\blacksquare$} other-ground} 
		& \rotatebox{90}{\textcolor{building}{$\blacksquare$} building} 
		& \rotatebox{90}{\textcolor{car}{$\blacksquare$} car} 
		& \rotatebox{90}{\textcolor{truck}{$\blacksquare$} truck} 
		& \rotatebox{90}{\textcolor{bicycle}{$\blacksquare$} bicycle} 
		& \rotatebox{90}{\textcolor{motorcycle}{$\blacksquare$} motorcycle} 
		& \rotatebox{90}{\textcolor{other-vehicle}{$\blacksquare$} other-vehicles} 
		& \rotatebox{90}{\textcolor{vegetation}{$\blacksquare$} vegetation} 
		& \rotatebox{90}{\textcolor{trunk}{$\blacksquare$} trunk} 
		& \rotatebox{90}{\textcolor{terrain}{$\blacksquare$} terrain} 
		& \rotatebox{90}{\textcolor{person}{$\blacksquare$} person} 
		& \rotatebox{90}{\textcolor{bicyclist}{$\blacksquare$} bicyclist}
		& \rotatebox{90}{\textcolor{motorcyclist}{$\blacksquare$} motorcyclist} 
		& \rotatebox{90}{\textcolor{fence}{$\blacksquare$} fence} 
		& \rotatebox{90}{\textcolor{pole}{$\blacksquare$} pole} 
		& \rotatebox{90}{\textcolor{traffic-sign}{$\blacksquare$} traffic-sign} \\
		\midrule
		
		$\checkmark$ & $\times$ & $\checkmark$ / $\times$ & $\times$ / $\times$ & 57.44 & 82.15 & 65.64 & 20.90 & 72.79 & 43.36 & 24.98 & 2.26 & 39.92 & 41.77 & 17.91 & 0.00 & 0.00 & 4.98 & 41.15 & 15.02 & \textbf{49.68} & 0.00 & 0.00 & 0.00 & 14.01 & 25.89 & 3.37 \\
		$\checkmark$ & $\times$ & $\checkmark$ / $\checkmark$ & $\times$ / $\times$ & 56.43 & 79.17 & 66.28 & 22.17 & 72.13 & 43.55 & \textbf{31.09} & 2.10 & 39.87 & 45.22 & 21.35 & 6.50 & 4.55 & 13.90 & 40.36 & 17.55 & 48.27 & 3.76 & 0.00 & 0.00 & 13.21 & 14.09 & 3.72 \\
		$\checkmark$ & $\checkmark$ & $\checkmark$ / $\times$  & $\checkmark$ / $\checkmark$ & 58.30 & \textbf{82.78} & 66.34 & 22.75 & 72.94 & 43.93 & 23.94 & 2.71 & 40.70 & 44.48 & 19.28 & 4.07 & 2.83 & 9.66 & \textbf{42.46} & 21.24 & 47.34 & 2.04 & 0.72 & 0.00 & \textbf{17.39} & \textbf{28.02} & 8.48 \\
		$\checkmark$ & $\checkmark$ & $\checkmark$ / $\times$ & $\times$ / $\times$ & \textbf{58.61} & 79.64 & 68.94 & 22.88 & \textbf{73.31} & 43.50 & 23.33 & 2.67 & 40.05 & 44.91 & 25.57 & 3.18 & 3.93 & 9.62 & 41.78 & 18.15 & 49.22 & 1.99 & 0.33 & 0.00 & 15.06 & 27.95 & \textbf{10.10} \\
		$\checkmark$ & $\checkmark$ & $\checkmark$ / $\checkmark$ & $\times$ / $\times$ & 57.90 & 80.35 & 67.45 & 23.96 & 73.11 & 43.70 & 24.27 & 2.85 & 41.09 & 46.74 & 29.54 & 7.93 & \textbf{8.09} & \textbf{19.84} & 41.62 & 21.86 & 49.77 & 5.85 & 1.34 & 0.00 & 14.14 & 18.24 & 5.26 \\
		$\checkmark$ & $\checkmark$ & $\checkmark$ / $\checkmark$ & $\checkmark$ / $\checkmark$ & 58.25 & 78.49 & \textbf{69.31} & \textbf{24.54} & 72.81 & \textbf{44.31} & 21.09 & \textbf{4.10} & \textbf{41.48} & \textbf{46.97} & \textbf{39.65} & \textbf{9.18} & 7.41 & 19.10 & 41.86 & \textbf{21.98} & 49.45 & \textbf{6.32} & \textbf{3.17} & 0.00 & 15.20 & 17.78 & 4.40 \\
		\bottomrule
	\end{tabular}\\
	\label{table:abalation}
\end{table*}

\section{Experiments}
\subsection{Dataset and Metrics}

\textbf{Dataset.} We test our method on the public semantic scene completion benchmark SemanticKITTI~\cite{behley2019semantickitti}. SemanticKITTI is a large-scale LiDAR point cloud dataset with a point-wise annotation collected by a single Velodyne HDL-64E laser scanner. The ground truth semantic labels of the scene completion task are composed of multiple consecutive point cloud frames with annotations. Following the official protocol, we select a single frame of the raw point cloud in the range of $[0 \sim 51.2 m, -25.6 \sim 25.6 m, -2 \sim 4.4 m]$ as input, and divide it by 0.2m to obtain voxels with a resolution of $256 \times 256 \times 32$. The output is the completed scene in the same area. The dataset contains 22 sequences with 19 semantic categories for training and testing. We use Sequences 0-7, 9-10 (3834 scans) for training, Sequence 8 (815 scans) for validation, and Sequences 11-21 (3901 scans) for testing.

\textbf{Metrics.}
We follow the regulations set by Song et al.~\cite{song2017semantic} to calculate IoU for scene completion, representing the completion of the scene (not involving semantics), and mIoU for semantic scene completion to measure the semantic segmentation performance over 19 classes of a completed scene. The metrics for semantic segmentation measurement mIoU is defined as:
\begin{equation}
   mIoU = \frac{1}{C} \sum_{c=1}^{C} \frac{{TP}_{c}}{{TP}_{c}+{FP}_{c}+{FN}_{c}}
\end{equation}
where ${TP}_{c}$, ${FP}_{c}$, ${FN}_{c}$ denote the number of true positive, false positive, and false negative predictions for class $c$ respectively, and $C$ denotes the number of classes.

\subsection{Implementation Details}

We augment the point cloud during training by randomly x-y flipping. We adopt the Adam optimizer~\cite{kingma2014adam} with a learning rate of 0.001 ($\beta_1 = 0.9$, $\beta_2 = 0.999$) for training, and each epoch is reduced by 2\%. All experiments are conducted on a single Nvidia GTX 1080 Ti with 11GB memory with batch size 2. 

\begin{figure*}[t]
	\centering
	\includegraphics[width=17.5cm]{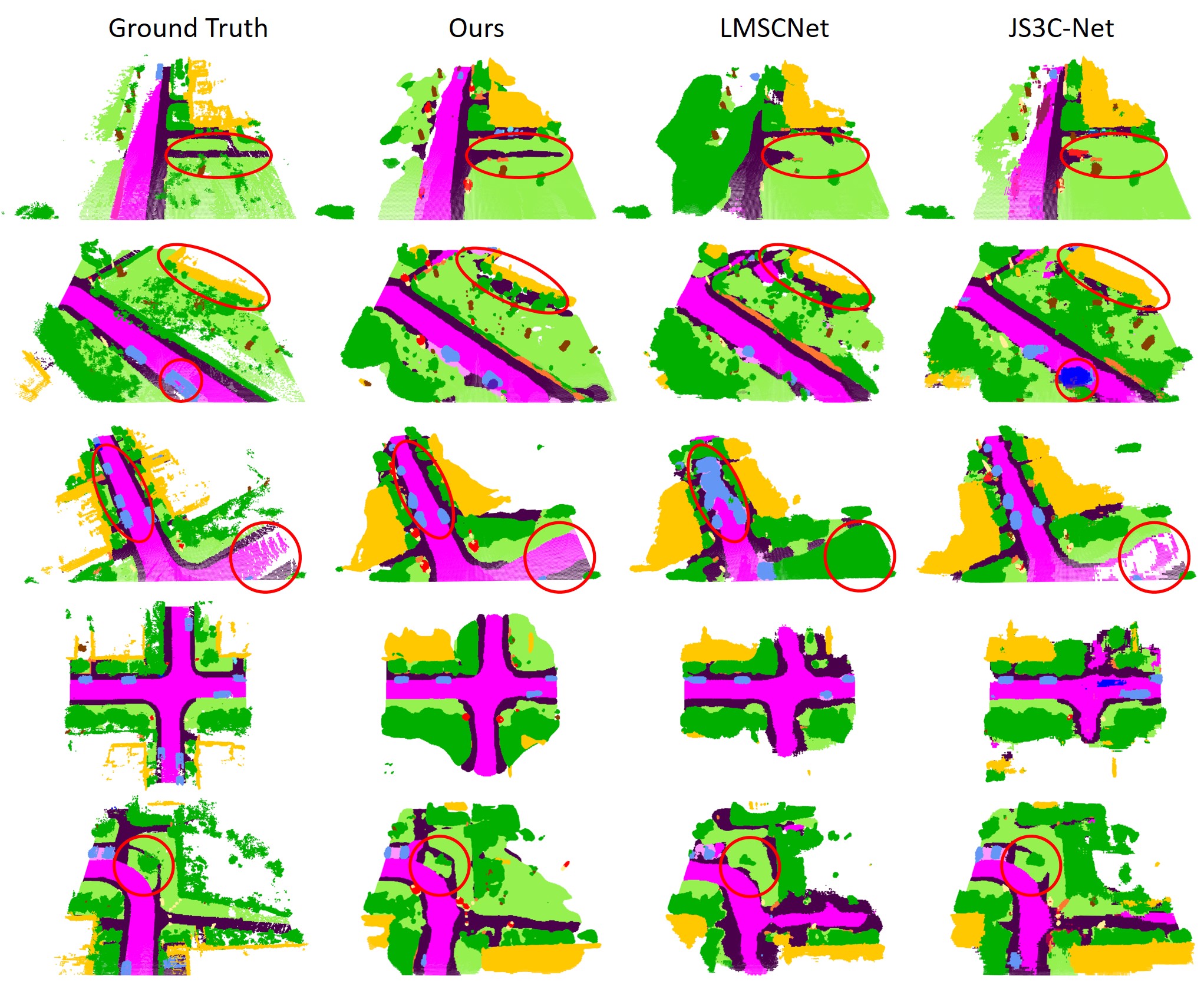}
	\caption{Comparison of qualitative results with other recent works. Experiments conducted on SemanticKITTI validation set.}
	\label{fig:results}
 \end{figure*}

\subsection{Quantitative Results and Analysis}

In this experiment, we submit the results of our work to the official evaluation server. The results of our method and the state-of-the-art methods are shown in Table.~\ref{table:net_reults}. It is worth noting that our method surpasses all previously published works in terms of completion metrics (IoU). By the time of submission, our method ranks $2^{nd}$ in completion metrics (IoU) and $3^{rd}$ in semantic completion metrics (mIoU) on the SemanticKITTI benchmark. In addition, we report the single scan inference latency on the entire validation split. Our network can reach a speed of 20.04 FPS with 2629 MB GPU memory when batch size is 1, which is much faster than other methods with comparable performance. In Table.~\ref{table:net_reults}, the numbers in brackets are measured on our device with batch size 1 using the official code.

Compared with the recently published voxel-based method LMSCNet~\cite{roldao2020lmscnet}, we have better results due to the integration of point cloud information. We have advantages on both IoU and mIoU, especially for mIoU, we have increased by 38.2\%. Compared with other integrated point cloud methods, we still have some advantages such as inference time. JS3C-Net~\cite{yan2020sparse} performs semantic segmentation on the point cloud before scene completion. Although it has obtained good mIoU results, it cannot achieve real-time results in inference time. S3CNet~\cite{cheng2020s3cnet} has achieved outstanding results in semantic scene completion, thanks to the geometric-aware loss which makes it have a very good performance on small objects such as \textit{bicycle} and \textit{motorcycle}. Nevertheless, it is not satisfied in terms of completion performance and inference time. Our method has better performance on ``plane'' categories, such as \textit{road} and \textit{sidewalk}. It is guessed that the 2D completion network has a better feature extraction effect on these categories and can better expand these features to the surroundings voxels.

\subsection{Qualitative Results}

We use the pretrained model and the official code of LMSCNet\footnote{\url{https://github.com/cv-rits/LMSCNet}} and JS3C-Net\footnote{\url{https://github.com/yanx27/JS3C-Net}}, and visualize the results in comparison with our results on SemanticKITTI validation set in Fig.~\ref{fig:results}. It can be seen that we do have obvious advantages in predicting the ``plane'' categories, which also verifies the metrics in Tab.~\ref{table:net_reults}. For some difficult samples, the results obtained by other methods are somewhat distorted, but our results can still get recognizable scenes. In addition, in terms of the completion performance, we can also make up a relatively complete scene better than other methods.

\subsection{Ablation Study}

In this part, we perform ablation studies on each part of the network to verify the effectiveness of the proposed method. The experiment results are shown in Tab.~\ref{table:abalation}. All experiments are trained on the training set, and the results are evaluated on the validation set.

The input for all experiments is point clouds. It can be seen that the baseline of our 2D UNet has already achieved good performance, especially for the ``plane'' categories which indicate that the 2D network is indeed good at feature diffusion. The addition of lovasz loss on this basis has further improved segmentation performance, because it can directly optimize mIoU, though it is not helpful for the completion metrics. Adding an assistant 3D branch on the baseline (no matter supervised by both lovasz loss and cross entropy loss or not) the network retains the ability to segment ``plane'' categories and also has a better effect on the segmentation of small objects. At the same time, the 3D structure helps to improve the performance of the completion. Finally, the semantic segmentation branch and lovasz loss are added on the baseline together, and the best results are obtained. Compared with the method without segmentation loss supervision, the proposed method obtains better performance because the semantic segmentation branch can provide semantic features instead of only a 3D structure. Furthermore, lovasz loss is also used to optimize the segmentation performance, they can promote each other with the semantic feature extraction and propagation. The lovasz loss allows the network to further improve the effect of small objects and accelerate the convergence, but at the same time, it seems to cause a decline on the completion metrics. 

\section{CONCLUSIONS}

In this paper, we propose a novel network for scene completion benefits from both 2D and 3D networks. The efficient 2D branch is used for completion. The 3D segmentation branch based on sparse convolution is to provide semantic features for the completion branch, bringing improvements in both completion and segmentation. The decoder of the segmentation branch can be discarded during inference, which will not increase too much computational burden. We also use lovasz loss to improve the network effect and accelerate convergence. Finally, our experiments show that the proposed model has real-time inference speed, and set a state-of-the-art performance on the SemanticKITTI dataset.

\addtolength{\textheight}{-19.5cm}   



\bibliographystyle{IEEEtran}
\bibliography{mybibfile}

\end{document}